\documentclass{article}
\usepackage{spconf,amsmath,graphicx}
\usepackage{booktabs}
\usepackage{comment}
\usepackage{amssymb} 
\usepackage{color}
\usepackage{caption}
\usepackage{booktabs, makecell}
\usepackage{colortbl}
\usepackage{siunitx}
\usepackage{multirow}
\usepackage{tabularx}
\usepackage{enumitem}
\usepackage{verbatim}
\usepackage{hyperref}
\usepackage{array}
\usepackage{booktabs}
\usepackage{subcaption}
\usepackage{float}
\usepackage{tcolorbox}
\usepackage[compact]{titlesec}
\usepackage{etoolbox}

\titlespacing{\section}{0pt}{0.8ex}{0.4ex}
\titlespacing{\subsection}{0pt}{0.6ex}{0.3ex}
\titlespacing{\subsubsection}{0pt}{0.5ex}{0.2ex}
\titleformat{\section}[block]%
  {\centering\normalfont\bfseries} 
  {\thesection.}
  {1em}%
  {}%

\titleformat{\subsection}[block]
  {\normalfont\normalsize\bfseries}
  {\thesubsection.}{1em}{}
\titlespacing{\subsection}{0pt}{0.6ex}{0.4ex}

\newtcolorbox{mybox}[1][]{colback=blue!3!white, colframe=pink!75!white, fonttitle=\bfseries, title=#1}

\title{Astrophotography Turbulence Mitigation via Generative Models}
\name{Joonyeoup Kim, Yu Yuan, Xingguang Zhang, Xijun Wang, Stanley Chan%
    \thanks{The work is supported, in part, by the National Science Foundation under the grants 2133032, 2030570, and Angular Encoded Imaging.}}

\address{School of Electrical and Computer Engineering, Purdue University}

\begin{document}

\maketitle

\begin{abstract}
Photography is the cornerstone of modern astronomical and space research. However, most astronomical images captured by ground-based telescopes suffer from atmospheric turbulence, resulting in degraded imaging quality. While multi-frame strategies like lucky imaging can mitigate some effects, they involve intensive data acquisition and complex manual processing. In this paper, we propose AstroDiff, a generative restoration method that leverages both the high-quality generative priors and restoration capabilities of diffusion models to mitigate atmospheric turbulence. Extensive experiments demonstrate that AstroDiff outperforms existing state-of-the-art learning-based methods in astronomical image turbulence mitigation, providing higher perceptual quality and better structural fidelity under severe turbulence conditions.
\end{abstract}

\begin{keywords}
Astronomical Image Processing, Turbulence Mitigation, Generative Models.
\end{keywords}

\section{INTRODUCTION}
\noindent
Atmospheric turbulence is a phenomenon rooted from random fluctuations in the Earth's atmosphere. They are primarily due to changes in temperature and pressure, which cause the refractive index of the air to change \cite{Jin2021NeutralizingTI,Jaiswal_2023_ICCV,zhang2023imagingatmosphereusingturbulence,zhang2024spatiotemporalturbulencemitigationtranslational,Takato:95}. These fluctuations induce both blurring from high-frequency light scattering and image jitter from large-scale wavefront tilts, significantly degrading astronomical observations. 
Many existing frameworks attempt to mitigate such effects through a phenomenon known as lucky imaging \cite{2008A&A...480..589B, rainer}. Lucky imaging employs high-frame-rate cameras to capture a large number of images, selects the sharpest \textcolor{black}{frames/regions}, and combines them to produce a high-\textcolor{black}{quality} image. While successful in certain scenarios, traditional methods have significant limitations, such as the need for image registration, low processing efficiency, noise accumulation, and potential information loss. In recent years, an increasing number of studies have utilized deep learning to address turbulence issues \cite{zhang2023imagingatmosphereusingturbulence, zhang2024spatiotemporalturbulencemitigationtranslational,9960930,SU202246}. However, few of these models are specifically tailored for astrophotography, resulting in suboptimal performance in astronomical image processing. These limitations stem not only from a lack of effective supervised training data but also from network architectures and training paradigms that fail to leverage high-quality prior information. To address these challenges, we harness the powerful generative priors and image restoration capabilities of generative models to mitigate turbulence in astronomical imaging.
Our work makes two key contributions:

\begin{enumerate}[leftmargin=*,noitemsep,nolistsep]
    \item \textbf{Datasets}: We present three rigorous datasets. \begin{itemize}
    \item \textbf{PlanetSYN}: 22,512 high-resolution astronomical images with dark backgrounds, combining real astronomical observations and synthetic textures.  
    \item \textbf{TechSYN}: 110,745 diverse images merging celestial features (sphere structures, astronomical textures) with terrestrial objects for generalized restoration learning.  
    \item \textbf{AstroEVA}: Evaluation benchmark containing 1,907 real telescope captures and turbulence-simulated images with ground-truth pairs.
\end{itemize}
    
    \item \textbf{Dual-branch Framework}: We present a novel architecture integrating a generative prior for structural preservation with a restoration model fused through SGLD
    
\end{enumerate}

\section{RELATED WORK}

\subsection{Lucky Imaging for Astrophotography Turbulence Mitigation}
\noindent
Lucky imaging is a well-established astrophotography technique that leverages high-speed cameras to capture thousands of short-exposure frames, taking advantage of brief moments when atmospheric turbulence is minimal \cite{2008A&A...480..589B, rainer, tpami}. During these optimal intervals, a subset of frames achieves near-diffraction-limited quality. These selected frames are then precisely aligned and merged to suppress noise and enhance the signal-to-noise ratio. 
Moreover, recent studies, like the work by Mao et al. \cite{9216531} or Joshi et al. \cite{rainer} have advanced lucky imaging to a greater degree by making it more efficient and even suitable for complex settings. 
Despite its success in delivering high-quality results in some cases, lucky imaging faces notable drawbacks. Its high frame rate requirement, usually above 40 fps leads to larger data volumes. For instance, the Nordic Optical Telescope reports that it captures about 1GB of data for 8 seconds, resulting in over 3 Terabytes of data volume overnight \cite{NOTLuckyCam}. 
Moreover, the process often involves additional processing steps, such as correcting for camera rotation, alignment of frames, and manually setting the proportion of lucky frames \cite{rainer}, making the overall workflow heavily dependent on manual configurations and labor-intensive, raising a need for a more automated data-driven alternative.

\subsection{Generative Models-based Image Enhancement}
\noindent
\textcolor{black}{Generative models, such as VAEs \cite{Kingma_2019}, GANs \cite{goodfellow2014generativeadversarialnetworks}, and diffusion models \cite{Sohl-Dickstein_2015_Nonequilibrium, ho2020denoisingdiffusionprobabilisticmodels}, are capable of capturing complex data distributions after being trained on large-scale datasets, enabling them to generate high-frequency details and textures that closely approximate real images.}
Recently, generative models have been applied to image restoration tasks, achieving remarkable success in areas including super-resolution \cite{saharia2021image}, denoising \cite{kawar2022denoisingdiffusionrestorationmodels}, deblurring \cite{ren2023multiscale}, and turbulence removal \cite{Jaiswal_2023_ICCV}. Among these, conditional diffusion models stand out by taking low-quality degraded images as guidance during the iterative denoising process, progressively refining them into high-quality restored samples. 

Despite these successes, generative models also have notable drawbacks. They can hallucinate details not present in the ground truth, and only a few address the unique textures, shapes, and extremely low-light conditions of astrophotography. Standard training data and architectures often fail to capture astronomical features like circular discs or ring formations, causing purely data-driven networks to struggle under severe turbulence and real telescope captures. 
But what if we could offset these limitations and maximize their full potential? 

\begin{mybox}[Question]
Can a domain-specific generative branch for astronomical images be fused with a specialized turbulence removal branch to outperform the typical data-driven deterministic methods, especially under extreme atmospheric conditions?
\end{mybox}


\section{METHODS}
\noindent
In this section, we will elaborate on the theoretical foundations, the constructed datasets, as well as the designed network with training methodology.

\subsection{Preliminary}
\noindent
In the following sections, we present a unique network inspired by Bayesian diffusion models \cite{xu2024bayesiandiffusionmodels3d} that integrates a generative prior branch with a restoration branch tailored for turbulence mitigation.

Recent studies in diffusion models have revolutionized image synthesis by iteratively transforming random noise into coherent images through a learned denoising process \cite{ho2020denoisingdiffusionprobabilisticmodels}. However, these models are prone to create unnecessary/unrealistic artifacts that did not exist in the original scene.

Bayesian Diffusion Models \cite{xu2024bayesiandiffusionmodels3d} address this challenge by combining a conditional likelihood term $P_{\gamma} (y | x)$ with an unconditional prior $P (y)$, dictated by the equation:
\setlength{\abovedisplayshortskip}{5pt}
\setlength{\belowdisplayshortskip}{5pt}
\begin{equation}
    p (y | x) \propto p_{\gamma} (y | x) * p(y)
\end{equation}
where x is a distorted image and y is the latent clean image. Through inference methods like Stochastic Gradient Langevin Dynamics \cite{Welling2011BayesianLV}, we can sample from the joint distribution explained above and successfully retrieve clean-cut images. 
By combining both the bottom-up structure model and the top-down structure model, this model architecture ensures that each output results in a clean image while maintaining structural integrity. 
However, there is a major challenge for this framework to be applied as a turbulence mitigation tool for astrophotography. Astronomical images are typically characterized by low signal-to-noise ratios, high contrast, and distinct structural figures. Moreover, atmospheric turbulence degrades these images in complicated ways that traditional restoration models cannot handle. The scarcity of high-quality data only worsens these challenges, as standard datasets do not capture the noise and distortion patterns unique to celestial observations. 
Our work addresses these issues by constructing large custom astronomical datasets and developing a dual-branch architecture. 


\subsection{Dataset}
\noindent
We constructed three different custom datasets to train and test our model.


\begin{figure} [H]
    \centering
    \includegraphics[width=1\linewidth]{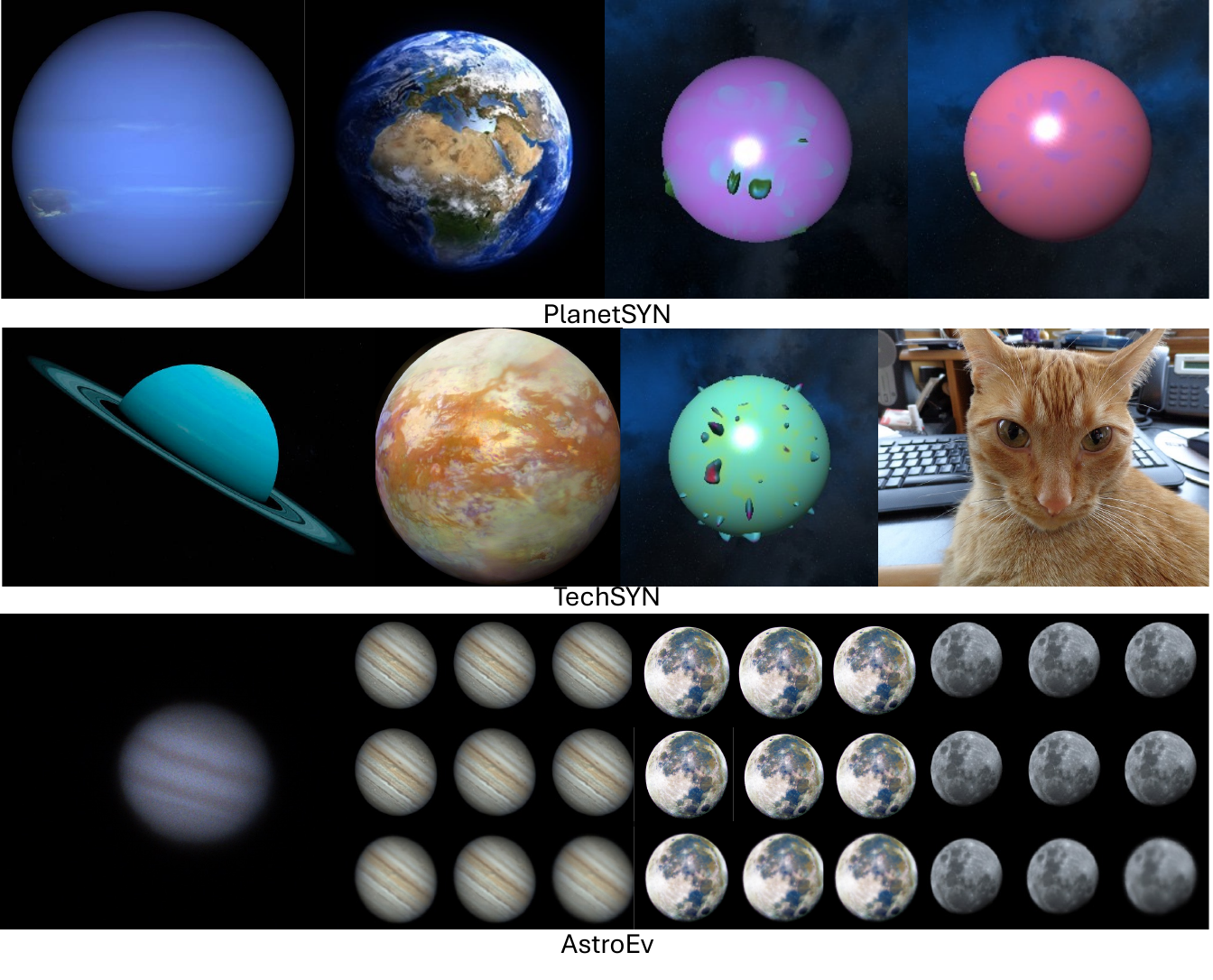}
    \caption{Sample images from the PlanetSYN dataset (top row), TechSYN dataset (second row), and AstroEVA dataset (bottom row)}
    \label{dataset}
\end{figure}

\begin{figure*}[tb]
    \centering
    \includegraphics[width=1\linewidth]{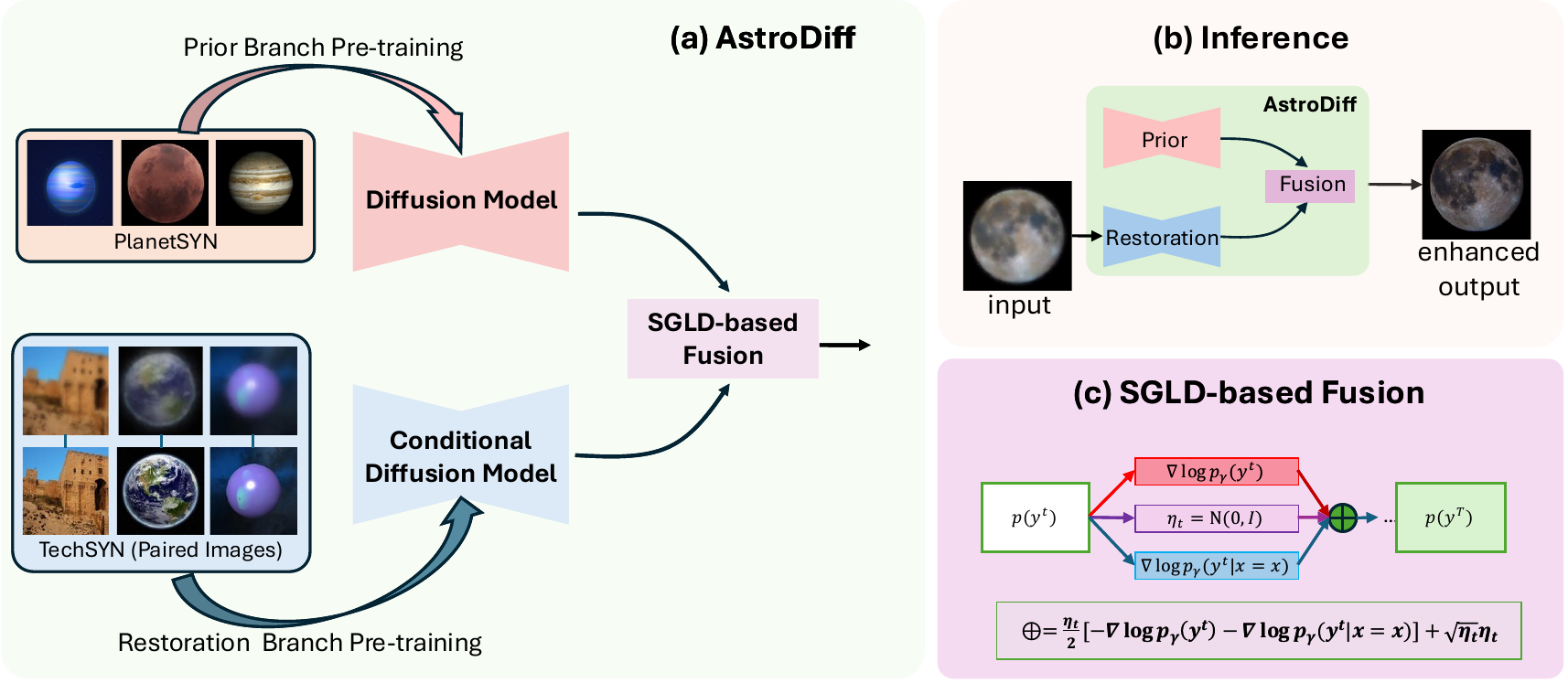}
    \caption{Overall Workflow of (a) AstroDiff - starts off with prior branch training focused on the generation of clean astronomical images, and restoration branch training focused on general turbulence mitigation, followed by inference phase shown in (b); (c) Detailed process of Stochastic Gradient Langevin Dynamics (SGLD) used to fuse the two branches.}
    \label{fig: workflow}
\end{figure*}

\textbf{PlanetSYN:} 
PlanetSYN was used to train the generative prior branch. It was constructed with 22,512 images. 2,609 images of different planets, ranging from Earth to Mercury, were collected from astrosurf.com. The rest of the dataset consisted of planet-like images collected from kaggle.com. As you can see from Figure \ref{dataset}, all images had a dark background with a sphere structure and planet-like textures. All of the images were then normalized and edited so that it would only contain necessary information.

\textbf{TechSYN:} 
TechSYN was created to train the restoration branch and was composed of 110,745 images. 60,000 images were collected from Kaggle.com, all of which had a dark background with a planet-like structure, just like the ones collected for the prior branch. 2,609 astronomical images that were used in the generative prior branch's training process were reused. The rest of the dataset is composed of random everyday images, ranging from a bike to an airplane, collected from ATSyn Static dataset \cite{zhang2024spatiotemporalturbulencemitigationtranslational} as shown in Figure \ref{dataset}. 

\textbf{AstroEVA:} 
AstroEVA, a custom dataset designed for model evaluation, is made up of 1,903 Jupiter images captured with a Celestron NexStar 127 SLT telescope and ZWO ASI462MC camera, an image of Mars captured with an Iphone 12, along with three astronomical images (Jupiter, Mercury, Moon - distinct from PlanetSYN or TechSYN) collected from astrosurf.com. We applied the noise simulator presented by Chan et al. \cite{zhang2024spatiotemporalturbulencemitigationtranslational} onto those five images with nine different $C_n^2$ values (from $5 \times 10^{-16}$ to $3 \times 10^{-13}$) to mimic turbulence.
We then used the BRISQUE metric \cite{Mittal2012NoReferenceIQ} to categorize the image sets into weak (0–40), medium (40–80), or strong (80+) turbulence levels, as it is shown in Table \ref{table:turbulence}.

\begin{table}
    \centering
    \resizebox{\columnwidth}{!}{
        \small 
        \renewcommand{\arraystretch}{1} 
        \setlength{\tabcolsep}{5pt}       
        \begin{tabular}{p{0.3\columnwidth} p{0.3\columnwidth} p{0.3\columnwidth}}
            \toprule
            \textbf{Low Turb} & \textbf{Medium Turb} & \textbf{High Turb} \\ 
            \textbf{Subject/$C_n^2$} & \textbf{Subject/$C_n^2$} & \textbf{Subject/$C_n^2$} \\ 
            \midrule
            Mercury/$5 \times 10^{-16}$   & Mercury/$5 \times 10^{-14}$      & Mercury/$1 \times 10^{-13}$    \\
            Mercury/$5 \times 10^{-15}$   & Moon/$5 \times 10^{-16}$         & Mercury/$2 \times 10^{-13}$    \\
            Mercury/$7 \times 10^{-15}$   & Moon/$5 \times 10^{-15}$         & Mercury/$3 \times 10^{-13}$    \\
            Mercury/$8 \times 10^{-15}$   & Moon/$7 \times 10^{-15}$         & Moon/$1 \times 10^{-13}$       \\
            Mercury/$1 \times 10^{-14}$   & Moon/$8 \times 10^{-15}$         & Moon/$2 \times 10^{-13}$       \\
            Jupiter/$5 \times 10^{-16}$   & Moon/$1 \times 10^{-14}$         & Moon/$3 \times 10^{-13}$       \\
            Jupiter/$1 \times 10^{-14}$   & Moon/$5 \times 10^{-14}$         & Jupiter/$1 \times 10^{-13}$    \\
            Jupiter/$5 \times 10^{-14}$   & Jupiter/$5 \times 10^{-15}$      & Jupiter/$2 \times 10^{-13}$    \\
                                          & Jupiter/$7 \times 10^{-15}$      & Jupiter/$3 \times 10^{-13}$    \\
                                          & Jupiter/$8 \times 10^{-15}$      &                                 \\
            \bottomrule 
        \end{tabular}
    }
    \caption{Comparison Table of Turbulence Levels}
    \label{table:turbulence}
\end{table}

\subsection{Proposed Method}

\noindent
The overall workflow of the model is shown in Figure \ref{fig: workflow}. Our training process is divided into two stages. In the first stage, we perform extensive pre-training separately for the generative prior branch and the restoration branch. In the second stage, we combine these two models through a fusion module using Stochastic Gradient Langevin Dynamics \cite{Welling2011BayesianLV}.

\begin{figure} [H]

    \centering
    \includegraphics[width=0.55\linewidth]{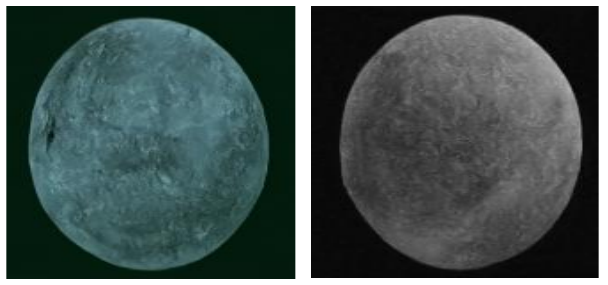}
    \caption{Sample images of synthetic astronomical images created by the generative prior branch}
    \label{fig:synthetic}
\end{figure}

\noindent \textbf{Stage 1.1: Prior Branch Pre-training:}
Our prior branch is trained solely on high-quality astronomical images, enabling a generative understanding of celestial objects. We use a U-Net diffusion approach \cite{ronneberger2015unetconvolutionalnetworksbiomedical}: each training iteration samples a random timestep $t$, adds Gaussian noise to a clean image, and trains the model to predict that noise, capturing typical astronomical features. After training, it was evaluated by generating synthetic images shown in Figure \ref{fig:synthetic}.

\begin{figure} [H]
    \centering
    \includegraphics[width=1\linewidth]{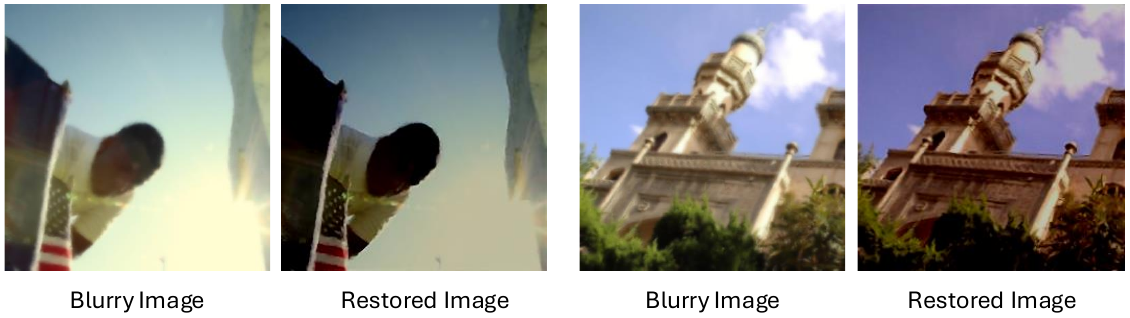}
    \caption{Samples of noisy image (first and third column) vs restored image using the restoration branch (second and last column)} 
    \label{fig:noise}
\end{figure}

\noindent \textbf{Stage 1.2: Restoration Branch Pre-training:}
The restoration branch focuses on turbulence mitigation by conditioning on distorted images and ground-truth pairs. It follows the same U-Net diffusion framework but learns to predict noise given a clean–and–noisy image pair at each timestep. Over multiple iterations, the model corrects turbulence-induced distortions in a bottom-up (data-driven) manner, as illustrated in Figure \ref{fig:noise}.


\noindent \textbf{Stage 2: SGLD-based Fusion:}
During fusion, we leverage a Bayesian Diffusion framework \cite{xu2024bayesiandiffusionmodels3d} that fuses bottom-up restoration and top-down generative models via Stochastic Gradient Langevin Dynamics (SGLD) \cite{Welling2011BayesianLV}. Starting from an initial guess of a clean image $y^{0}$, each iteration refines the image by:
\begin{enumerate}
    \item \textbf{Random Timestep Selection:} Sample a diffusion timestep $t$, setting the expected noise level.
    \vspace{-0.5em}
    \item \textbf{Gradient Computation:} 
    \begin{itemize}[leftmargin=*,noitemsep,nolistsep]
        \item \emph{Data-Driven Gradient:} Derived from the  restoration branch to ensure $y^{t}$ aligns with the observed frame.
        \item \emph{Generative Gradient:} Derived from the prior branch to steer $y^{t}$ towards plausible astronomical features.
    \end{itemize}
    \vspace{-0.5em}  
    \item \textbf{Fusion and Update:} Blend the gradients with fixed weights, update $y^{i}$, and add controlled Gaussian noise to avoid local minima, using the equation:
    \begin{equation}
    \resizebox{0.9\linewidth}{!}{$
        y^{t+1}={\frac{\eta_t}{2}}[-\nabla{log{\rho}_{\gamma}(y^t) - \nabla{log{\rho}_{\gamma}(y^t|x = x)}]+ \sqrt{{\eta}_t}\eta_t}
    $}
    \end{equation}
    where ${\eta}_t \sim \mathcal{N}(0, I)$
    \vspace{-0.5em} 
    \item \textbf{Refinement:} Iterate until $y^{t+1}$ transitions from noise to a coherent reconstruction.
\end{enumerate}

\section{\textbf{EXPERIMENTS}}

\begin{figure} [H]
    \centering
    \includegraphics[width=1\linewidth]{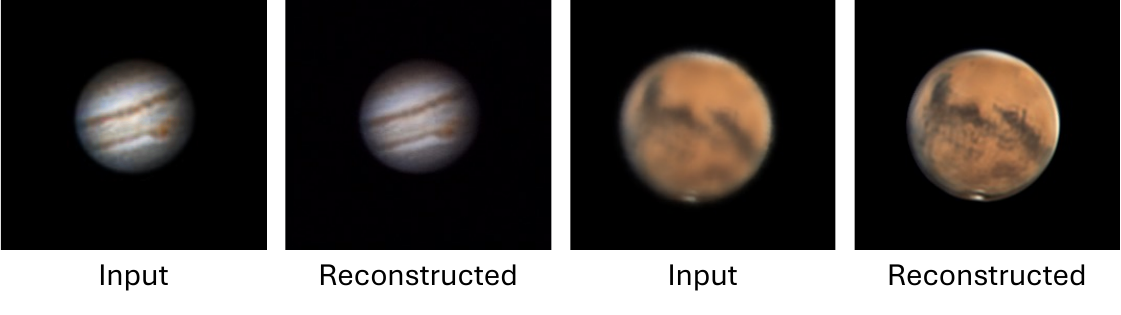}
    \caption{Sample images of Mars and Jupiter in AstroEVA (first column) vs enhanced Image of Mars and Jupiter using AstroDiff (second column)}
    \label{fig:jupiter}
\end{figure}

\begin{table} [H]
    \centering
    \renewcommand{\arraystretch}{1}
    \setlength{\tabcolsep}{12pt}
    \begin{tabular}{l c}
        \toprule
        \textbf{Method} & \textbf{BRISQUE} $\downarrow$ \\
        \midrule
        Original Images     & 95.64 \\
        Enhanced Images & \colorbox{blue!10}{\textbf{22.78}} \\
        \bottomrule
    \end{tabular}
    \caption{BRISQUE evaluation on real data (lower is better). 
             Reconstruction achieves \textbf{76.2\%} improvement.}
    \label{tab:brisque_results}
\end{table}

\subsection{Implementation Details}
\noindent
We adopted a U-Net architecture \cite{ronneberger2015unetconvolutionalnetworksbiomedical} processing $256 \times 256$ images for both branches. The models were trained with a batch size of 8 using the AdamW optimizer (learning rate = \(10^{-4}\), weight decay = \(10^{-6}\)). A cosine annealing scheduler with a 5\% warmup was employed over 50,000 training steps to ensure stable convergence.

\subsection{Evaluation on Real Data}
\noindent
We evaluated our model's performance on real data in the AstroEVA dataset. The model was evaluated on the average BRISQUE \cite{Mittal2012NoReferenceIQ} score of 1,907 images. Table \ref{tab:brisque_results} and Figure \ref{fig:jupiter} denote AstroDiff's successful turbulence mitigation results.

\begin{figure*}[tb]
    \centering
    \includegraphics[width=1\linewidth]{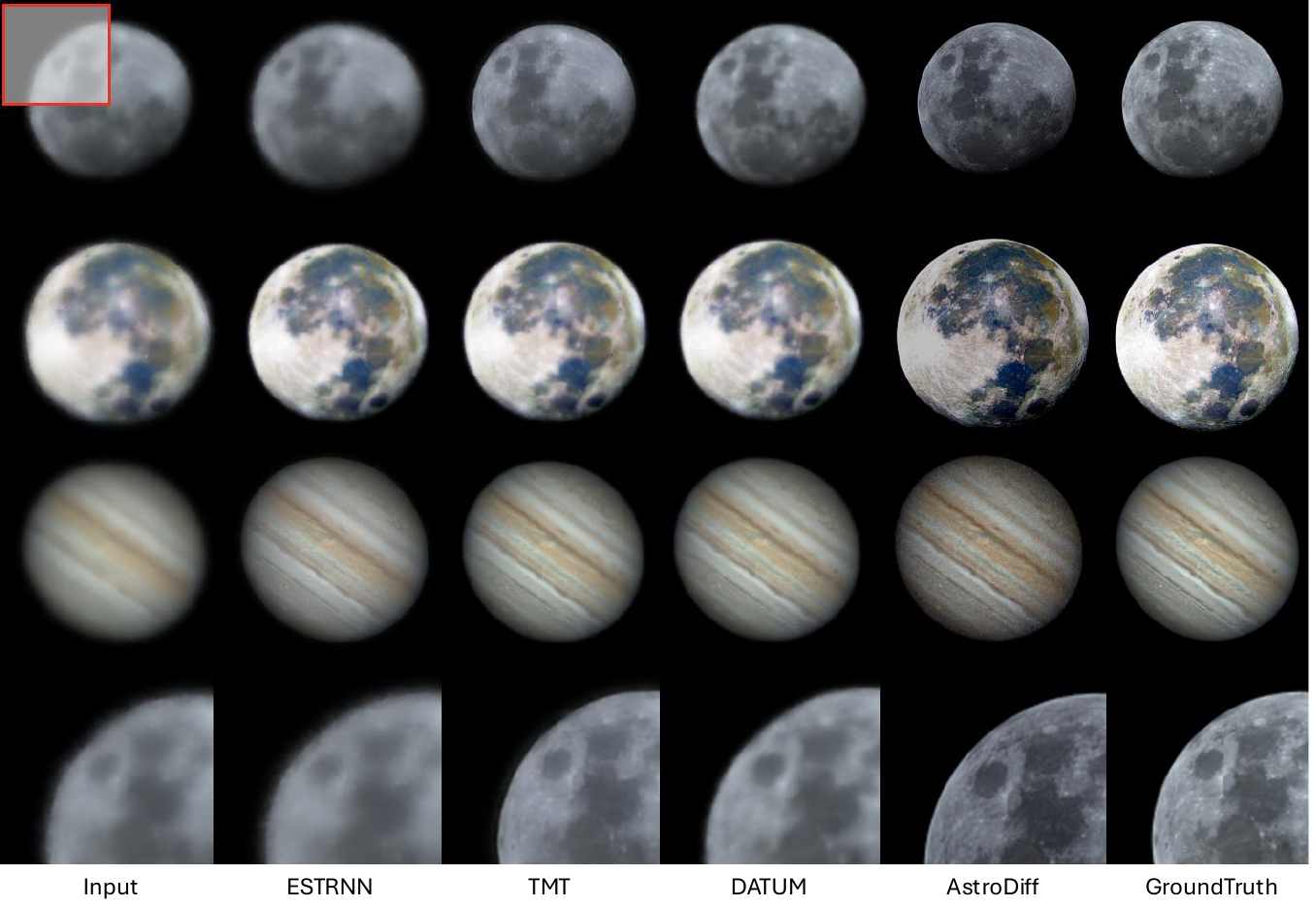}
    \caption{Qualitative comparison of AstroEVA Dataset Under High Turbulence ($C_n^2 = 3 \times 10^{-13}$) - last row indicates the zoomed-in area dictated by the red box in the first row.}
    \label{baseline_compl}
\end{figure*}

\subsection{Baselines and Metrics}
\noindent
Three state-of-the-art turbulence mitigation models (DATUM \cite{zhang2024spatiotemporalturbulencemitigationtranslational}, TMT \cite{zhang2023imagingatmosphereusingturbulence}, ESTRNN \cite{zhong2022realworldvideodeblurringbenchmark}) were selected as baselines.
The training scheme for each model was the same as the original paper. 
For each set, the models were evaluated on their performance based on the average PSNR and LPIPS \cite{zhang2018unreasonableeffectivenessdeepfeatures} values over the rest of the AstroEVA dataset.

\begin{table}[H]
    \centering
    \resizebox{\columnwidth}{!}{
    \renewcommand{\arraystretch}{1}
    \setlength{\tabcolsep}{4pt}
    \begin{tabular}{l *{2}{c} *{2}{c} *{2}{c}}
        \toprule
        \multirow{2}{*}{Method} & 
        \multicolumn{2}{c}{Low Turb} & 
        \multicolumn{2}{c}{Medium Turb} & 
        \multicolumn{2}{c}{High Turb} \\
        \cmidrule(lr){2-3} \cmidrule(lr){4-5} \cmidrule(lr){6-7}
        & PSNR $\uparrow$ & LPIPS $\downarrow$ & PSNR $\uparrow$ & LPIPS $\downarrow$ & PSNR $\uparrow$ & LPIPS $\downarrow$ \\
        \midrule
        DATUM \cite{zhang2024spatiotemporalturbulencemitigationtranslational}
          & \cellcolor{blue!10}\textbf{31.09} & 0.11
          & \cellcolor{blue!10}\textbf{32.12} & 0.09
          & 30.44 & 0.12 \\
        TMT \cite{zhang2023imagingatmosphereusingturbulence}
          & 29.57 & 0.13
          & 30.03 & 0.13
          & 30.49 & 0.11 \\
        ESTRNN \cite{zhong2022realworldvideodeblurringbenchmark}
          & 29.02 & 0.18
          & 29.58 & 0.16
          & 28.83 & 0.14 \\
        AstroDiff (ours)
          & 29.29 & \cellcolor{blue!10}\textbf{0.07}
          & 30.22 & \cellcolor{blue!10}\textbf{0.08}
          & \cellcolor{blue!10}\textbf{30.86} & \cellcolor{blue!10}\textbf{0.07} \\
        \bottomrule
    \end{tabular}
    }
    \caption{Quantitative comparison of turbulence mitigation across different levels. The best results are highlighted and shown in bold.}
    \label{table:baselinecomp}
\end{table}

\subsection{Comparisons} 
\noindent
Although Figure \ref{baseline_compl} shows attractive images, Table \ref{table:baselinecomp} reports relatively low PSNR, which is a known trade-off in generative models that favor perceptual quality over pixel-perfect accuracy. This is reflected in our superior LPIPS scores, as our network preserves high-level structures rather than matching pixels. Another crucial advantage of our model emerges under severe turbulence, where conventional pipelines \cite{zhang2024spatiotemporalturbulencemitigationtranslational}, \cite{zhang2023imagingatmosphereusingturbulence}, \cite{zhong2022realworldvideodeblurringbenchmark},
\cite{9216531} often generate artifacts or overly smoothed reconstructions. In contrast, AstroDiff's dual branch design enforces both restoration and generative constraints: the learned prior steers outputs toward physically plausible structures, while the restoration branch ensures adherence to the clean-data distribution. As a result, even in extreme turbulence, our model retains global coherence unlike traditional methods.





\subsection{Ablation Study}
\noindent
To further evaluate our two-model architecture, we compared the average PSNR and LPIPS values of the restoration branch versus the final dual-branch model on the same baseline images.

\begin{figure} [H]
    \centering
    \includegraphics[width=1\linewidth]{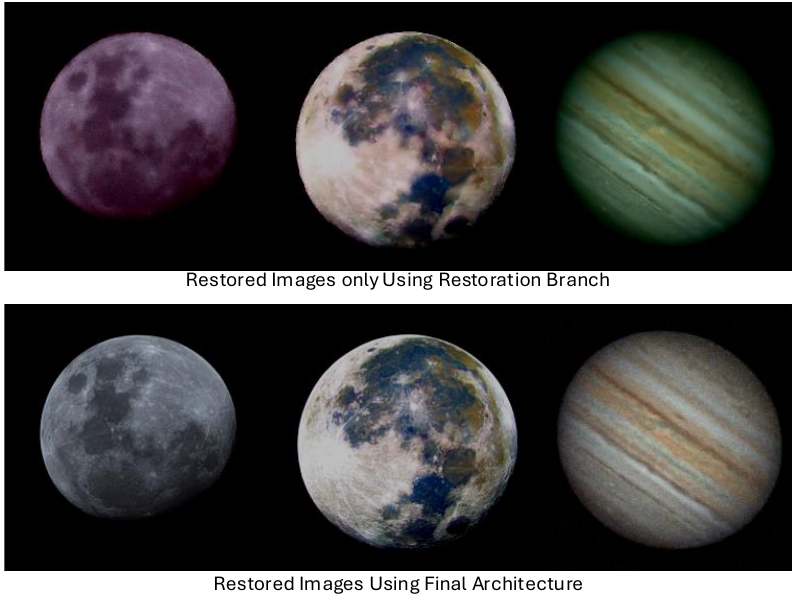}
    \caption{Enhanced images of AstroEVA generated exclusively with the restoration branch (top) compared to enhanced images of AstroEVA produced using the final model (bottom).}
    \label{fig:comp1}
\end{figure}

As shown in Figure \ref{fig:comp1} and Table \ref{table:merged_metrics}, not only does AstroDiff create a more visually pleasing image, but it also clearly outperforms the one-step model in every metric.

\begin{table}[H]
    \centering
    \resizebox{\columnwidth}{!}{
        \small
        \renewcommand{\arraystretch}{1.2}
        \setlength{\tabcolsep}{4pt}
        \begin{tabular}{l*{6}{c}}
            \toprule
            \multirow{2}{*}{Method} & 
            \multicolumn{2}{c}{Low Turb} & 
            \multicolumn{2}{c}{Medium Turb} & 
            \multicolumn{2}{c}{High Turb} \\
            \cmidrule(lr){2-3}\cmidrule(lr){4-5}\cmidrule(lr){6-7}
            & PSNR $\uparrow$ & LPIPS $\downarrow$ 
            & PSNR $\uparrow$ & LPIPS $\downarrow$
            & PSNR $\uparrow$ & LPIPS $\downarrow$ \\
            \midrule
            One Step Model  
            & 20.54 & 0.18
            & 26.65 & 0.19
            & 24.42 & 0.17 \\
            
            Two Step Model  
            & \cellcolor{blue!10}\textbf{29.29} & \cellcolor{blue!10}\textbf{0.07} 
            & \cellcolor{blue!10}\textbf{30.02} & \cellcolor{blue!10}\textbf{0.08} 
            & \cellcolor{blue!10}\textbf{29.86} & \cellcolor{blue!10}\textbf{0.07} \\
            \bottomrule
        \end{tabular}
    }
    \caption{Quantitative comparison for ablation study.}
    \label{table:merged_metrics}
\end{table}

\section{\textbf{CONCLUSION}}
\noindent
In this paper, We present AstroDiff, a new network designed for astrophotography. AstroDiff is built by combining a generative prior branch and a restoration branch, and fused with SGLD inference. Experimental results demonstrate effective atmospheric turbulence removal and structural preservation of our architecture, making it the best turbulence mitigation model for astrophotography. 
\clearpage

{
\bibliographystyle{IEEEtran} 
\bibliography{bibliography}

\begin{thebibliography}{10}
\providecommand{\url}[1]{#1}
\csname url@samestyle\endcsname
\providecommand{\newblock}{\relax}
\providecommand{\bibinfo}[2]{#2}
\providecommand{\BIBentrySTDinterwordspacing}{\spaceskip=0pt\relax}
\providecommand{\BIBentryALTinterwordstretchfactor}{4}
\providecommand{\BIBentryALTinterwordspacing}{\spaceskip=\fontdimen2\font plus
\BIBentryALTinterwordstretchfactor\fontdimen3\font minus
  \fontdimen4\font\relax}
\providecommand{\BIBforeignlanguage}[2]{{%
\expandafter\ifx\csname l@#1\endcsname\relax
\typeout{** WARNING: IEEEtran.bst: No hyphenation pattern has been}%
\typeout{** loaded for the language `#1'. Using the pattern for}%
\typeout{** the default language instead.}%
\else
\language=\csname l@#1\endcsname
\fi
#2}}
\providecommand{\BIBdecl}{\relax}
\BIBdecl

\bibitem{Jin2021NeutralizingTI}
D.~Jin, Y.~Chen, Y.~Lu, J.~Chen, P.~Wang, Z.~Liu, S.~Guo, and X.~Bai,
  ``Neutralizing the impact of atmospheric turbulence on complex scene imaging
  via deep learning,'' \emph{Nature Machine Intelligence}, vol.~3, pp. 876 --
  884, 2021.

\bibitem{Jaiswal_2023_ICCV}
A.~Jaiswal, X.~Zhang, S.~H. Chan, and Z.~Wang, ``Physics-driven turbulence
  image restoration with stochastic refinement,'' in \emph{Proceedings of the
  IEEE/CVF International Conference on Computer Vision (ICCV)}, October 2023,
  pp. 12\,170--12\,181.

\bibitem{zhang2023imagingatmosphereusingturbulence}
X.~Zhang, Z.~Mao, N.~Chimitt, and S.~H. Chan, ``Imaging through the atmosphere
  using turbulence mitigation transformer,'' \emph{arXiv preprint arXiv:
  2207.06465}, 2023.

\bibitem{zhang2024spatiotemporalturbulencemitigationtranslational}
X.~Zhang, N.~Chimitt, Y.~Chi, Z.~Mao, and S.~H. Chan, ``Spatio-temporal
  turbulence mitigation: A translational perspective,'' \emph{arXiv preprint
  arXiv: 2401.04244}, 2024.

\bibitem{Takato:95}
N.~Takato and I.~Yamaguchi, ``Spatial correlation of zernike phase-expansion
  coefficients for atmospheric turbulence with finite outer scale,'' \emph{J.
  Opt. Soc. Am. A}, vol.~12, no.~5, pp. 958--963, May 1995.

\bibitem{2008A&A...480..589B}
J.~E. {Baldwin}, P.~J. {Warner}, and C.~D. {Mackay}, ``The point spread
  function in lucky imaging and variations in seeing on short timescales,''
  \emph{\aap}, vol. 480, no.~2, pp. 589--597, Mar. 2008.

\bibitem{rainer}
N.~Joshi and M.~Cohen, ``Seeing mt. rainier: Lucky imaging for multi-image
  denoising, sharpening, and haze removal,'' in \emph{2010 IEEE International
  Conference on Computational Photography (ICCP)}, 2010.

\bibitem{9960930}
A.~Christopher, R.~Hari~Kishan, and P.~Sudeep, \emph{Image Reconstruction Using
  Deep Learning}.\hskip 1em plus 0.5em minus 0.4em\relax John Wiley \& Sons,
  Inc., 2023, pp. 65--87.

\bibitem{SU202246}
J.~Su, B.~Xu, and H.~Yin, ``A survey of deep learning approaches to image
  restoration,'' \emph{Neurocomputing}, vol. 487, pp. 46--65, 2022.

\bibitem{tpami}
X.~Zhu and P.~Milanfar, ``Removing atmospheric turbulence via space-invariant
  deconvolution,'' \emph{IEEE Transactions on Pattern Analysis and Machine
  Intelligence}, vol.~35, no.~1, pp. 157--170, 2013.

\bibitem{9216531}
Z.~Mao, N.~Chimitt, and S.~H. Chan, ``Image reconstruction of static and
  dynamic scenes through anisoplanatic turbulence,'' \emph{IEEE Transactions on
  Computational Imaging}, vol.~6, pp. 1415--1428, 2020.

\bibitem{NOTLuckyCam}
``Luckycam,'' \url{https://www.not.iac.es/instruments/luckycam/}, accessed:
  February 05, 2025.

\bibitem{Kingma_2019}
D.~P. Kingma and M.~Welling, ``An introduction to variational autoencoders,''
  \emph{Foundations and Trends® in Machine Learning}, vol.~12, no.~4, p.
  307–392, 2019.

\bibitem{goodfellow2014generativeadversarialnetworks}
I.~J. Goodfellow, J.~Pouget-Abadie, M.~Mirza, B.~Xu, D.~Warde-Farley, S.~Ozair,
  A.~Courville, and Y.~Bengio, ``Generative adversarial networks,'' \emph{arXiv
  preprint arXiv: 1406.2661}, 2014.

\bibitem{Sohl-Dickstein_2015_Nonequilibrium}
J.~Sohl-Dickstein, E.~Weiss, N.~Maheswaranathan, and S.~Ganguli, ``Deep
  unsupervised learning using nonequilibrium thermodynamics,'' in
  \emph{Proceedings of the 32nd International Conference on Machine Learning},
  vol.~37, 2015, pp. 2256--2265.

\bibitem{ho2020denoisingdiffusionprobabilisticmodels}
J.~Ho, A.~Jain, and P.~Abbeel, ``Denoising diffusion probabilistic models,''
  \emph{arXiv preprint arXiv: 2006.11239}, 2020.

\bibitem{saharia2021image}
C.~Saharia, J.~Ho, W.~Chan, T.~Salimans, D.~J. Fleet, and M.~Norouzi, ``Image
  super-resolution via iterative refinement,'' \emph{arXiv preprint arXiv:
  2104.07636}, 2021.

\bibitem{kawar2022denoisingdiffusionrestorationmodels}
B.~Kawar, M.~Elad, S.~Ermon, and J.~Song, ``Denoising diffusion restoration
  models,'' \emph{arXiv preprint arXiv: 2201.11793}, 2022.

\bibitem{ren2023multiscale}
M.~Ren, M.~Delbracio, H.~Talebi, G.~Gerig, and P.~Milanfar, ``Multiscale
  structure guided diffusion for image deblurring,'' in \emph{Proceedings of
  the IEEE/CVF International Conference on Computer Vision}, 2023, pp.
  10\,721--10\,733.

\bibitem{xu2024bayesiandiffusionmodels3d}
H.~Xu, Y.~Lei, Z.~Chen, X.~Zhang, Y.~Zhao, Y.~Wang, and Z.~Tu, ``Bayesian
  diffusion models for 3d shape reconstruction,'' \emph{arXiv preprint arXiv:
  2403.06973}, 2024.

\bibitem{Welling2011BayesianLV}
M.~Welling and Y.~W. Teh, ``Bayesian learning via stochastic gradient langevin
  dynamics,'' in \emph{International Conference on Machine Learning}, 2011.

\bibitem{Mittal2012NoReferenceIQ}
A.~Mittal, A.~K. Moorthy, and A.~C. Bovik, ``No-reference image quality
  assessment in the spatial domain,'' \emph{IEEE Transactions on Image
  Processing}, vol.~21, pp. 4695--4708, 2012.

\bibitem{ronneberger2015unetconvolutionalnetworksbiomedical}
O.~Ronneberger, P.~Fischer, and T.~Brox, ``U-{N}et: Convolutional networks for
  biomedical image segmentation,'' \emph{arXiv preprint arXiv: 1505.04597},
  2015.

\bibitem{zhong2022realworldvideodeblurringbenchmark}
Z.~Zhong, Y.~Gao, Y.~Zheng, B.~Zheng, and I.~Sato, ``Real-world video
  deblurring: A benchmark dataset and an efficient recurrent neural network,''
  \emph{arXiv preprint arXiv: 2106.16028}, 2022.

\bibitem{zhang2018unreasonableeffectivenessdeepfeatures}
R.~Zhang, P.~Isola, A.~A. Efros, E.~Shechtman, and O.~Wang, ``The unreasonable
  effectiveness of deep features as a perceptual metric,'' \emph{arXiv preprint
  arXiv: 1801.03924}, 2018.

\end{thebibliography}
}

\end{document}